\documentclass{article} % For LaTeX2e
\usepackage{iclr2015,times}
\usepackage{hyperref}
\usepackage{bm}
\usepackage{subfig}
\usepackage{graphicx,epstopdf,times,amsmath,amssymb}
\DeclareGraphicsExtensions{.pdf,.eps,.png,.jpg,.mps, bmp}

\title{ Classifier with Hierarchical Topographical Maps as Internal Representation}

\author{
Thomas Trappenberg \& Paul Hollensen \\
Faculty of Computer Science \\
Dalhousie University \\
Halifax, Canada \\
\texttt{\{tt,hollensen\}@cs.dal.edu}
\And
Pitoyo Hartono \\
School of Engineering\\
Chukyo University\\
Nagoya, Japan \\
\texttt{hartono@ieee.org} \\
}

\iclrfinalcopy % Uncomment for camera-ready version

\begin{document}

\maketitle

\begin{abstract}
In this study we want to connect our previously proposed context-relevant topographical maps with the deep learning community. Our architecture is a classifier with hidden layers that are hierarchical two-dimensional topographical maps. These maps differ from the conventional self-organizing maps in that their organizations are influenced by the context of the data labels in a top-down manner. In this way bottom-up and top-down learning are combined in a biologically relevant representational learning setting. Compared to our previous work, we are here specifically elaborating the model in a more challenging setting compared to our previous experiments and to advance more hidden representation layers to bring our discussions into the context of deep representational learning.
 \end{abstract}

\section{Introduction}

Early cortical representations of sensory information in the mammalian brain have spatial organizations \citep{hubel1962receptive}. For example, simple cells in the primary visual cortex respond to edges at specific locations, while adjacent cells tend to respond best to edges with only small variations of orientations and locations. Hubel and Wiesel identified such hypercolumns in which edge orientations are topographically represented. Topographic representations are also present in other sensory modalities such as in the primary auditory cortex \citep{romani1982tonotopic} and the somatosensory cortex \citep{wall1983recovery}. Models to describe self-organizations of such sensory maps have long been studied \citep{willshaw1976patterned, Kohonen082}. %, ChemicalGuidance}
In the context of representational learning, such algorithms must be considered as unsupervised learning algorithms that are purely data driven, and we characterize such learning also as bottom-up learning.

In contrast, recent developments in deep learning have overwhelmingly shown the abilities of supervised learning in which representations are mainly guided by back-propagating errors from labelled data \citep{krizhevsky2012imagenet}. While early deep learning approaches have emphasized the advantage of unsupervised pre-training \citep{Hinton06}, most current successes do not rely on this form of unsupervised learning, which in their basic form do not include topographic constraints. %and furthermore these unsupervised methods do not appear to benefit from topographic constraints
It has hence become a major puzzle why biological system have this form of representational organization. In terms of learning theories, such organizations represent representational constraints that on a first glimpse can restrict the classification performance of deep learning classifiers. 

\begin{figure}[htbp]
  \begin{center}
	\includegraphics[width=6cm,height=6cm]{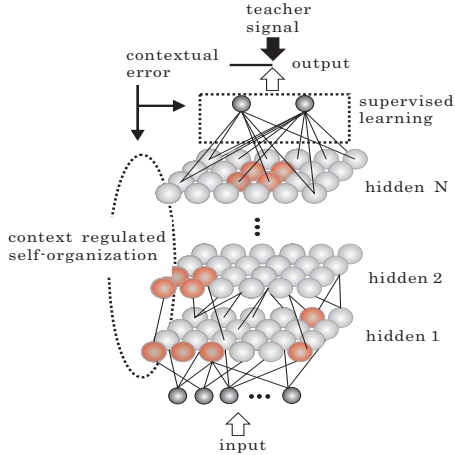}
        \caption{Multilayerd Restricted Radial Basis Function Network \label{fig:mrrbf}}
  \end{center}
\end{figure}

In this paper we study the interaction between bottom-up self-organization and top-down error correction learning in a simple model that combines a Kohonen-type self-organizing map (SOM) with backprogagation learning in some classification tasks. We have introduced such a model with a single hidden layer in \citet{Hartono014} and tested this on simple machine learning benchmarks.  Here we start to apply this model to more challenging tasks and to generalize our model to versions with multiple hidden layers, as outlined in Fig. \ref{fig:mrrbf}, in order to ultimately scale the model to much larger problems. The Kohonen SOM uses nodes that have Gaussian activation functions, and backpropagating errors through such a network corresponds to a radial basis function (RBF) network. However, the RBF network is restricted by the bottom-up constraints of the SOM, and thus we called this network a restricted RBF (rRBF). Interestingly, by combining the Kohonen SOM learning with backpropagation, we found that the top down error has the effect of enforcing the topographic neighbourhood of items with the same label while "repelling" representations of items with different labels.

The evaluation of the rRBF with one hidden SOM layer on basic machine learning examples has shown that there is a much stronger clustering of classes in the hidden layer compared to the standard SOM and that the classification performance is comparable to a basic mulitlayer perceptron. Hence, while we can argue that our approach aids in the visualization of the representations, our approach does not lead to better classifications that would excite machine learning practitioners. Including label information in SOM has also been done in various ways in the past \citep{Ritter089}. However, as mentioned before, our main aim here is to understand the consequences of biological facts -- here, topographic representations -- in terms of deep learning representations. More specifically, our initial tests have been simple examples where a shallow 2-dimensional representation is mostly sufficient. Here we ask if more complicated relationships can be represented with deeper version of this architecture, and how such a version would compare to traditional deep learning approaches. 

\section{The multilayered restricted radial basis function network}

The multilayered restricted radial basis function network (M-rRBF) contains an input layer, an output layer and a hierarchical stack of $N$ hidden layers between them as shown in Fig. \ref{fig:mrrbf}. 

The activation $I^M_k$ and corresponding $O^M_k$  of the $k$-th neuron in the $M$-th layer of the M-rRBF are that of a standard radial basis function network,
\begin{eqnarray}
I^M_k(t) &=& \frac{1}{2} \|\bm{W}^M_k(t) - \bm{O}^{M-1}(t)\|^2  \hspace{2 mm} \mbox {for}  \hspace{2 mm} (1 \leq M \leq N)  \nonumber \\
O^M_k(t) &=& e^{-I^M_k(t)} \sigma(k^{*M},k,t) \label{eq:output},
\end{eqnarray}
where $\bm{W}^M_k(t)$ is the reference vector associated with the $k$-th neuron in the $M$-th layer. Here, $\bm{O}^0(t) = \bm{X}(t)$ where  $\bm{X}(t)$ is the input vector at time $t$. In this equation, $k^{*M}$ is the best matching unit (BMU) in the $M$-th hidden layer,
\begin{equation}
w^{M} = arg \min_k I^M_k(t) \label{eq:win}.
\end{equation}
%The neighbourhood function $\sigma()$ is constantly decreasing with regard to the distance between the neuron $k$ and the BMU,
The neighbourhood function, $\sigma$, becomes narrower over the course of learning according to 
\begin{eqnarray}
\sigma(k^{*M},k,t) &=& e^{-\frac{\|k^{*M} - k \|^2}{s(t)}} \label{eq:neighbor} \\
s(t) &=& s_0 (\frac{s_{end}}{s_0})^{\frac{t}{t_{end}}} \nonumber
\end{eqnarray}
The output layer is a sigmoidal perceptron, 
\begin{eqnarray}
I_l(t) &=& \sum_{k} v_{kl}(t) O^N_k(t) - \theta_l(t) \\
y_l(t) &=& \frac{1}{1+e^{-I_l(t)}},
\label{eq:outputneuron}
\end{eqnarray}
where $v_{kl}(t)$ is the weight connecting the $k$-th neuron in the last hidden layer, and $\theta_l(t)$ is the bias of that neuron at time $t$.

We train the network on a quadratic error function at time $t$, 
\begin{equation}
E(t) =\frac{1}{2} \sum_l (y_l(t) - T_l(t))^2,
\label{eq:errorfunction}
\end{equation}
where $T_l(t)$ is the $l$-th component of the teacher signal at time $t$. The connection weights from the $N$-th hidden layer to the output layer and corresponding biases of the output neurons are minimized with gradient descent 
\begin{eqnarray}
v_{kl}(t+1) &=& v_{kl}(t) -\eta_1 \frac{\partial E(t)}{\partial v_{kl}(t)} \\
\theta_{l}(t+1) &=& \theta_{l}(t) -\eta_1 \frac{\partial E(t)}{\partial \theta_{l}(t)}.
\label{eq:modifyout}
\end{eqnarray}
The gradients of the output weights is given by
\begin{eqnarray}
\frac{\partial E(t)}{\partial v_{kl}(t)} &=& \frac{\partial E(t)}{\partial y_l(t)} \frac{\partial y_l(t)}{\partial I_l(t)} \frac{\partial I_l(t)}{\partial  v_{kl}(t)} \nonumber  \\
&=& \delta_l (t) O^N_k(t) \\
\delta_l(t) &=& (y_l(t) - T_l(t)) y_l(t) (1 - y_l(t)),
\end{eqnarray}
and similarly for the biases. The gradient descent of the reference vectors associated with the neurons in the $N$-th hidden layer is
\begin{equation}
W^N_{jk}(t+1) = W^N_{jk}(t) - \eta_{hid} \frac{\partial E(t)}{\partial W^N_{jk}} \label{eq:modifyN1} 
\end{equation}
where  $\eta_{hid}$ is the learning rate for the hidden layers. This leads to the learning rule
\begin{eqnarray}
W_{jk}^N(t+1) &=& W_{jk}^N(t) + \eta_{hid} \delta^N_k(t) \sigma(k^{*N},k,t) (O_j^{N-1}(t) - W_{jk}^N(t)) \label{eq:modifyN3} \\
\delta^N_k(t) &=& -(\sum_l \delta_l(t) v_{kl}(t)) e^{-I^N_k(t)} \label{eq:deltaN}.
\end{eqnarray}
If $\bm{O}^{N-1}$ is considered to be the input vector to the $N$-th hidden layer, the modification in Eq. \ref{eq:modifyN3} is similar to the modification rule of the conventional SOM. The only difference is that in SOM, the reference vector is always pulled toward the input vector, while in Eq. \ref{eq:modifyN3}, it is not necessarily so. The direction of the reference vector's modification is decided by the sign of $\delta_k^N(t)$, in which a positive $\delta_k^N(t)$ modifies the reference vector as in SOM, while a negative one repels the reference vector from the input vector. This  $\delta_k^N(t)$ is a regulatory signal of error feedback from the supervised layer, and thus reflects the label context to the otherwise purely self-organizing process in this layer.

The modification of the reference vectors in the $(N-L)$-th layer  with $(1 \leq L \leq N-1)$ is correspondingly given by \begin{eqnarray}
W^{N-L}_{ab}(t+1) &=& W^{N-L}_{ab}(t) + \eta_{hid} \Delta W_{ab}^{N-L}(t)  \label{eq:hierarchy1} \\
\Delta W_{ab}^{N-L}(t)  &=& \delta_b^{N-L}(t) \sigma(k^{*N-L},b,t) (O^{N-L-1}_a(t) - W_{ab}^{N-L}(t))  \label{eq:hierarchy2} \\
\delta_b^{N-L}(t) &=& (-1)^{l} (\sum_a \Delta W_{ab}^{N-L+1}(t)) e^{-I_b^{N-L}(t)}  \label{eq:hierarchy3}
\end{eqnarray}
The equations show that also in this deeper version of our architecture the organizing process in the $(N-L)$-th layer is regulated by contextual error, $\delta_b^{N-L}$, that is passed from the higher layer.

\section{Experiments}

\subsection{Difference between MLP, SOM and CRSOM}

It is often criticized that a MLP is a black box in which the operation at each level is not readily understood in terms of simple rules. A frequent motivating goal of SOMs is therefore the embeddings of a feature space in a low-dimensional, visualizable form. The primary difference between CRSOM and SOM is that SOM preserves the topological structures of high-dimensional data into a low dimensional map based on their similarities of the data alone, as often measured by Euclidean distance, while CRSOM also incorporates their context as provided by the class labels. It should be noted that similarities in the features of data do not always translate into the similarities of their contexts. For example the physical features of lion and zebra, such as their sizes, number of legs, place to live and running speed are very similar, so if they are characterized with those features they should be mapped closely in a SOM. However, if they are viewed in the context of carnivore-herbivore, they should be mapped far from each other.

\begin{figure}[htbp]
  \begin{center}
    \includegraphics[width=14cm]{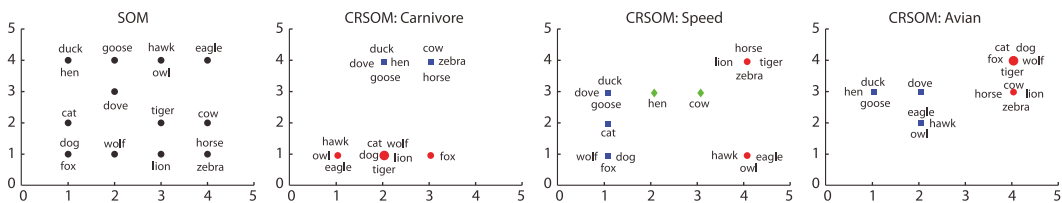}
    \caption{Mappings learned from the animals dataset \citet{Ritter089}. From left to right: SOM (no context), CRSOM with "carnivore" context, "speed" context, and "avian" context.} \label{fig:animalsom}
  \end{center}
\end{figure}

To demonstrate the difference between SOM and CRSOM we ran preliminary experiments using the animal data set proposed in \citet{Ritter089}. The data set in this experiment contains 16 animals, each one characterized by 16 binary features. Figure \ref{fig:animalsom} shows the mapping from the 16-dimensional data to a 2-dimensional map learned by conventional SOM and three CRSOMs provided different contexts. For example, the SOM maps "duck" and "hen", which share similar features, into a single point which is far from the point where "horse" is mapped to.

To illustrate the context-relevance of CRSOM, different contexts were infused into the data. The first CRSOM mapping in Figure \ref{fig:animalsom} (middle left) shows the CRSOM produced by training a single-layered rRBF in which the original animal data were labeled either as "carnivore" or "herbivore".  In this CRSOM, neurons that were chosen as winners for herbivores were drawn as $\square$, and neurons for carnivores were drawn as $\bigcirc$, while their sizes reflects their winning frequencies.  In this CRSOM only fives neurons were chosen as winners: two in the upper half of the map are occupied by herbivores while three neurons in the lower-half are occupied by carnivores. It is interesting to observe that the infusion of contexts changed the appearance of the low-dimensional representation of the data. For example in the original SOM "duck" and "zebra" were diagonally distanced from each other, but in the carnivore/herbivore-contexted CRSOM they are positioned close to each other due to their common context of "herbivore". It should be noted that CRSOM also preserves the topological characteristics of the data, for example "duck", "dove", "hen" and "goose" that share similar features are mapped into a single point.

In order to see how a shifting context would influence the hidden representations we ran two more experiments with altered class structure.  In the second experiment each data point was labeled according to the animals' moving velocity, either as "fast", "medium" or "slow". These are illustrated as $\bigcirc$, $\Diamond$ and $\square$, respectively, in Fig. \ref{fig:animalsom} (middle right). In the third experiment an rRBF was trained to classify the data points into either of avian or non-avian, and the resulting CRSOM is shown in Fig. \ref{fig:animalsom} (far right), where $\square$ illustrate a neuron for avians and $\bigcirc$ is a neuron for non-avians. The three experiments with different class context show that the CRSOM preserves the topographical characteristics of high dimensional data in relevance to their contexts.

\subsection{CRSOM on MNIST data}

\begin{figure}[htbp]
   \centering
{\includegraphics[width=13cm,height=4cm]{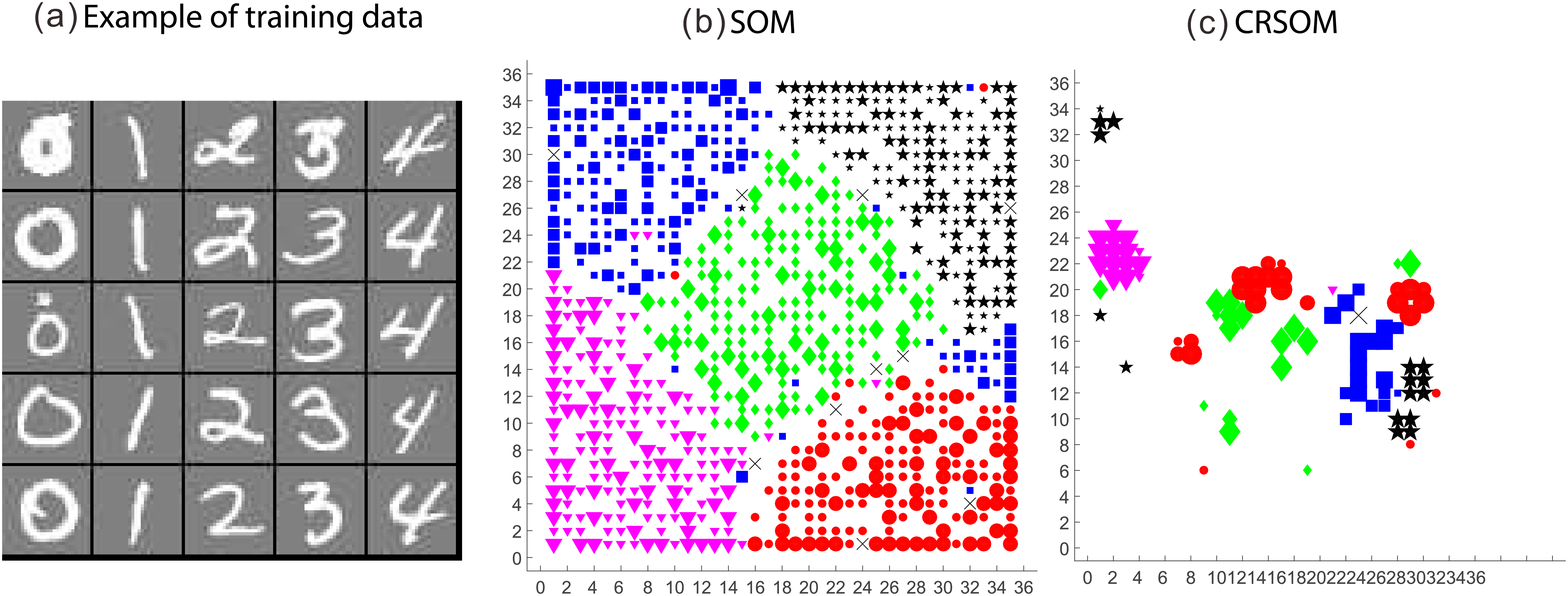}}\label{fig:mnistdata}
   \caption{MNIST database: 784 features, 5 classes, 1269 instances}\label{fig:mnist}
\end{figure}

To start comparing our biological motivated model with more typical deep learning benchmarks, we trained a single-layered rRBF with part of the MNIST database of handwritten digits. We only used the digits from 0 to 4 in the following illustrations mainly for time reasons as we did not have the time to experiment with larger networks.  However, this is not a principle restiction and larger models are now under investigation. Some of the training examples are shown in Fig. \ref{fig:mnist}(a).

The conventional SOM for part of the MNIST handwriting data is shown in Fig. \ref{fig:mnist}(b) and the CRSOM produced by the single-layered rRBF in this experiment is shown in Fig. \ref{fig:mnist}(c). In both figures, red dots indicate digit 4, blue is for 3, green is for 2, blue is for 1 and magenta is for 0. From these figures, we can observe that the representation of rRBF for this problem is different from the purely topographical characteristics of the data. The generated CRSOM is also sparser and provides more information than the conventional SOM. For example from Fig. \ref{fig:mnist} we can intuitively learn that the red dots (digit 4) are more varied than for example digit 0. To show that the CRSOM is a better representation than SOM, we also tested the classification performance of the rRBF with CRSOM as its internal layer, and a classifier with SOM as the internal layer. In this test, the network with CRSOM as its internal layer produced $0.24\%$ classification error, versus $1.5\%$ for the SOM-based network.

\subsection{First results on deeper representations}

In order to test how a second layer will influence the learned representations in our architecture, we repeated experiments on well known UCI datasets. While these experiments might not be exciting in terms of improving practical machine learning applications, our aim here is to understand how representational learning would look in a biological motivated network.

\begin{figure}[htbp]
     \centering
   {\includegraphics[width=13cm]{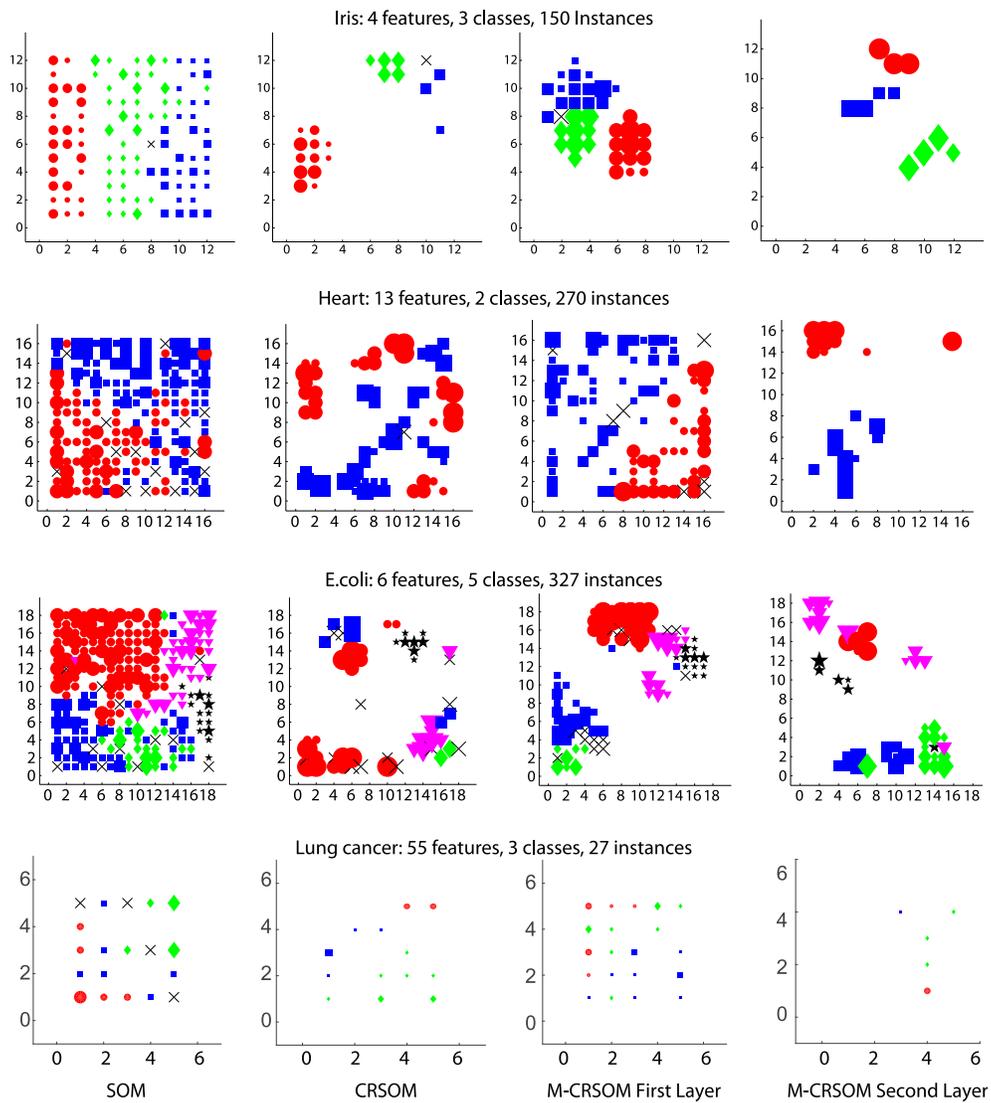}}\label{fig:irisetalsom}
     \caption{Mappings learned by SOM, CRSOM, and M-CRSOM on the UCI dataset.}\label{fig:irisetal}
\end{figure}

Figure \ref{fig:irisetal} shows the mappings learned by SOM and by CRSOM with 1 and 2 hidden layers on the Iris, Heart, E.Coli and Lung cancer datasets. For the SOM, the contexts (classes of the data) did not have any influence to the organization of the map, but for visualization clarity they were drawn with different shapes in the map, in which their sizes show their winning frequencies, and $\times$s show the reference vectors that were selected as winners for inputs belonging to different classes. Taking Iris as an example, the single layer CRSOM, as well as the first layer of the M-CRSOM, nicely illustrates the characteristics of this problem, where one class is easily separable, while the remaining two are not easily separable from each other.  The second layer of the CRSOM benefits from the depth of the preceeding representation in order to fully separate the classes.  The SOM mapping, on the other hand, does not offer any obvious information about the class structure.  Across the datasets, the CRSOM representation is consistently much sparser, and this sparsity increases with layer depth. Margins between class clusters also consistently increase with layer depth, and some datasets which were non-separable in the first layer become so in the second.

\begin{figure}[htbp]
  \begin{center}
    \includegraphics[width=12cm,height=5cm]{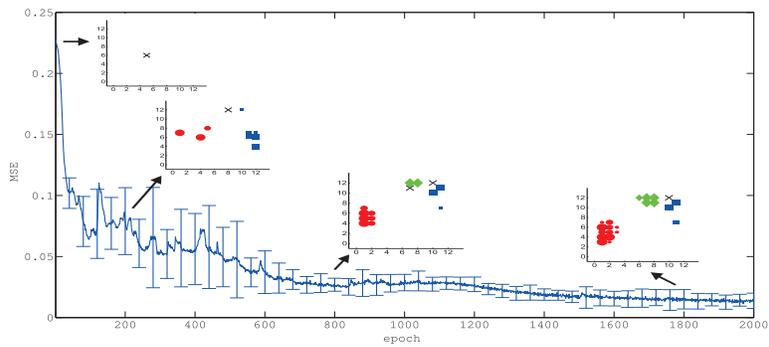}
    \caption{Learning Curve: average of learning error of Iris problem over 10 trials, shown with CRSOM formation during the learning process\label{fig:irisetallearn}}
  \end{center}
\end{figure}

Figure \ref{fig:irisetallearn} shows the average error during the learning process of Iris problem over 10 trials, with the typical gradual map formation process during the learning process.


\subsection{Generalization of rRBF}

\begin{figure}[htbp]
 \begin{center}
   \includegraphics[width=12cm, height=5cm]{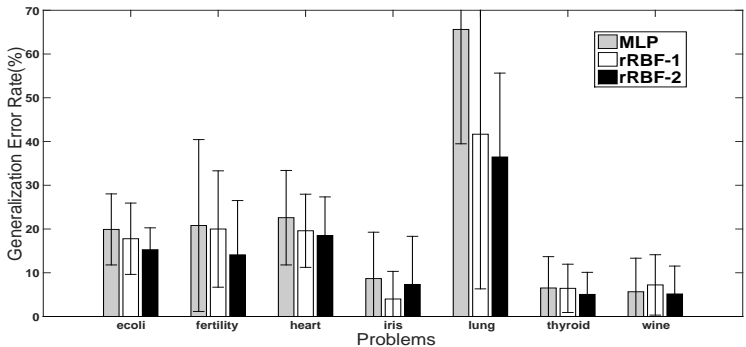}\label{fig:generror}
   \caption{Generalization Comparison}\label{fig:generalization}
  \end{center}
\end{figure}

The generalization performance of rRBF is tested with a 10-fold cross validation for some classification problems as shown in Fig. \ref{fig:generalization}. The performance of the classifiers are statistically equivalent for all the methods, given the standard deviations plotted as error bars. This is an encouraging result as our previous tests with single hidden layer networks did not consistently perform as well as comparators.

While at this point there is no statistically significant difference in the classification performance, average errors do decrease with increasing depth of representation.  It is also interesting to observe that the correlation between the generalization performance and the visual appearance of the map. The appearance of the CRSOM is correlated with the generalization performance of the rRBF. For example, in the Iris problems, where well- separated clusters were formed, the rRBF usually produced high generalization performance. Such clustered representations should also produce more robustness to representational noise.

\section{Discussion and conclusions}
This paper is intended to connect our previous work on context relevant maps and known biological representations with the deep learning community. We showed that a generalization of the formulas to architectures with many hidden layers is straight forward and still preserves the finding that context from class labels can bias the topographic maps in the manner found previously.

The first experiments with the deeper network are very encouraging. While this is still a simple problem, the results show that a much better clustering of the IRIS data are achieved relative to the single hidden layer case. It is clear that the influence of different contexts for data will make this a much more difficult clustering problem compared to the situation where only bottom-up self-organization is considered. It should hence not be expected that such representations, as well as representations of much more complex data, are achieved with simple 2-dimensional maps. The hierarchical architecture offers here a solution in that early representations could still reflect higher dimensional relations while higher representations could work more efficiently on the simplified and compressed previous representations.

While our preliminary tests are encouraging us to look further into applying this model to more complicated large scale benchmarks, it is still not clear what the advantage of topographic organization in biological systems is. It is clear that visualization can not be an argument for our brain, and in terms of machine learning it seems that topography is a hindering constraint. However, given that nature had time to optimize brain architectures, it is very possible that the topographic constraints allow better generalizations in problem domains that are specifically addressed by the brain.

\bibliography{hartonov1}

\begin{thebibliography}{9}
\providecommand{\natexlab}[1]{#1}
\providecommand{\url}[1]{\texttt{#1}}
\expandafter\ifx\csname urlstyle\endcsname\relax
  \providecommand{\doi}[1]{doi: #1}\else
  \providecommand{\doi}{doi: \begingroup \urlstyle{rm}\Url}\fi

\bibitem[Hartono et~al.(2014)Hartono, Hollensen, and Trappenberg]{Hartono014}
Hartono, Pitoyo, Hollensen, Paul, and Trappenberg, Thomas.
\newblock Learning-regulated context relevant topographic maps.
\newblock \emph{IEEE Trans. on Neural Networks and Intelligent Systems}, (in
  press), 2014.

\bibitem[Hinton et~al.(2006)Hinton, Osindero, and Teh]{Hinton06}
Hinton, Geoffrey~E., Osindero, Simon, and Teh, Yee~Whye.
\newblock A fast learning algorithm for deep belief nets.
\newblock \emph{Neural Computation}, 18:\penalty0 1527--1554, 2006.

\bibitem[Hubel \& Wiesel(1962)Hubel and Wiesel]{hubel1962receptive}
Hubel, David~H and Wiesel, Torsten~N.
\newblock Receptive fields, binocular interaction and functional architecture
  in the cat's visual cortex.
\newblock \emph{The Journal of physiology}, 160\penalty0 (1):\penalty0
  106--154, 1962.

\bibitem[Kohonen(1982)]{Kohonen082}
Kohonen, Teuvo.
\newblock Self-organized formation of topologically correct feature maps.
\newblock \emph{Biological Cybernetics}, 43:\penalty0 59--69, 1982.

\bibitem[Krizhevsky et~al.(2012)Krizhevsky, Sutskever, and
  Hinton]{krizhevsky2012imagenet}
Krizhevsky, Alex, Sutskever, Ilya, and Hinton, Geoffrey~E.
\newblock Imagenet classification with deep convolutional neural networks.
\newblock In \emph{Advances in neural information processing systems}, 2012.

\bibitem[Ritter \& Kohonen(1989)Ritter and Kohonen]{Ritter089}
Ritter, Helge and Kohonen, Teuvo.
\newblock A self-organizing semantic maps.
\newblock \emph{Biological Cybernetics}, 61:\penalty0 241--254, 1989.

\bibitem[Romani et~al.(1982)Romani, Williamson, and
  Kaufman]{romani1982tonotopic}
Romani, Gian~Luca, Williamson, Samuel~J, and Kaufman, Lloyd.
\newblock Tonotopic organization of the human auditory cortex.
\newblock \emph{Science}, 216\penalty0 (4552):\penalty0 1339--1340, 1982.

\bibitem[Wall et~al.(1983)Wall, Felleman, and Kaas]{wall1983recovery}
Wall, John~T, Felleman, Daniel~J, and Kaas, Jon~H.
\newblock Recovery of normal topography in the somatosensory cortex of monkeys
  after nerve crush and regeneration.
\newblock \emph{Science}, 221\penalty0 (4612):\penalty0 771--773, 1983.

\bibitem[Willshaw \& Von Der~Malsburg(1976)Willshaw and Von
  Der~Malsburg]{willshaw1976patterned}
Willshaw, David~J and Von Der~Malsburg, Christoph.
\newblock How patterned neural connections can be set up by self-organization.
\newblock \emph{Proceedings of the Royal Society of London B: Biological
  Sciences}, 194\penalty0 (1117):\penalty0 431--445, 1976.

\end{thebibliography}
\bibliographystyle{iclr2015}

\appendix

\section{Appendix}
%Similar to the modification of the last hidden layer, the modification of the reference vectors in the $(N-1)$-th layer can be calculated as follows.
The modification of the reference vectors in the $(N-1)$-th layer can be calculated as follows.

\begin{eqnarray}
\frac{\partial E(t)}{\partial W_{ij}^{N-1}(t)}  &=& \frac{\partial{E(t)}}{\partial y_l(t)} \frac{\partial y_l(t)}{\partial W_{ij}^{N-1}(t)} \nonumber \\
&=& \sum_l (y_l(t) - T_l(t)) y_l(t) (1-y_l(t)) \frac{\partial I_l(t)}{\partial W_{ij}^{N-1}(t)} \nonumber \\
&=& -\sum_k (\sum_l \delta_l(t) v_{kl}(t)) O^N_k(t) (O^{N-1}_j(t) - W_{jk}^N(t)) \frac{\partial O^{N-1}_j(t)}{\partial W_{ij}^{N-1}(t)} \nonumber \\
&=&(\sum_k \Delta W_{jk}^N(t)) e^{-I_j^{N-1}}(t) \sigma(k^{*N-1},j,t) (O^{N-2}_i(t) - W_{ij}^{N-1}(t)) \\
\Delta W_{jk}^N(t) &=& \delta_k^N(t) \sigma(k^{*N},k,t) (O_j^{N-1}(t) - W_{jk}^N(t)) \\
W_{ij}^{N-1}(t+1) &=& W_{ij}^{N-1}(t) + \eta_{hid} \delta_j^{N-1}(t) \sigma(k^{*N-1},j,t) (O^{N-2}_i(t) - W^{N-1}_{ij}(t)) \\
\delta_j^{N-1}(t) &=& -(\sum_k \Delta W^N_{jk}(t)) e^{-I_j^{N-1}(t)} 
\end{eqnarray}
%\end{eqnarray}
%
%Hence,
%
%\begin{eqnarray}
Following this chain rule,  the modification of the reference vectors in the $(N-L)$-th layer  $(1 \leq L \leq N-1)$ can be written as follows.

\begin{eqnarray}
W^{N-L}_{ab}(t+1) &=& W^{N-L}_{ab}(t) + \eta_{hid} \Delta W_{ab}^{N-L}(t)  \label{eq:hierarchy1_1} \\
\Delta W_{ab}^{N-L}(t)  &=& \delta_b^{N-L}(t) \sigma(k^{*N-L},b,t) (O^{N-L-1}_a(t) - W_{ab}^{N-L}(t))  \label{eq:hierarchy2_1} \\
\delta_b^{N-L}(t) &=& (-1)^{l} (\sum_a \Delta W_{ab}^{N-L+1}(t)) e^{-I_b^{N-L}(t)}  \label{eq:hierarchy3_1}
\end{eqnarray}

% \section{Appendix 2}
% 
% Figure \ref{fig:irisetallearn} shows the average error during the learning process of Iris problem over 10 trials, with the typical gradual map formation process during the learning process.
% \begin{figure}[htbp]
%   \begin{center}
%     \includegraphics[width=12cm,height=5cm]{./figures/iriscrsomlearn.eps}
%     \caption{Learning Curve: average of learning error of Iris problem over 10 trials, shown with CRSOM formation during the learning process\label{fig:irisetallearn}}
%   \end{center}
% \end{figure}

\end{document}